\documentclass{article}

\usepackage{microtype}
\usepackage{graphicx}
\usepackage{booktabs}
\usepackage{hyperref}
\usepackage{multirow}
\usepackage{xcolor}
\usepackage{enumitem}

\usepackage[accepted]{icml2026}

\makeatletter
\renewcommand{\ICML@appearing}{\textit{Workshop on Trustworthy AI for Good
at the 43${}^{\text{rd}}$ International Conference on Machine Learning},
Seoul, South Korea, 2026.
Copyright 2026 by the author(s).}
\makeatother

\usepackage{amsmath}
\usepackage{amssymb}
\usepackage{mathtools}

\usepackage[capitalize,noabbrev]{cleveref}

\icmltitlerunning{Susceptibility and Acknowledgment}

\begin{document}

\twocolumn[
\icmltitle{Beyond Accuracy: Measuring Bias Acknowledgment \\ in Chain-of-Thought Reasoning for Responsible AI Evaluation}

\icmlsetsymbol{equal}{*}

\begin{icmlauthorlist}
\icmlauthor{Xian Sun}{duke,equal}
\icmlauthor{Wei Gao}{neu,equal}
\icmlauthor{Yingshuo Wang}{berkeley}
\icmlauthor{Lingdong Kong}{nus}
\icmlauthor{Yanhang Li}{neu}
\icmlauthor{Zhichao Fan}{uiuc}
\icmlauthor{Zexin Zhuang}{smu}
\icmlauthor{Wenlong Dong}{uca}
\icmlauthor{Zhiyuan Zheng}{ind}
\icmlauthor{Hrishikesh Paranjape}{ind}
\icmlauthor{Abhishek Mandal}{ind}
\icmlauthor{Johnny R. Zhang}{ind}
\end{icmlauthorlist}

\icmlaffiliation{duke}{Duke University}
\icmlaffiliation{berkeley}{University of California, Berkeley}
\icmlaffiliation{neu}{Northeastern University}
\icmlaffiliation{nus}{National University of Singapore}
\icmlaffiliation{uiuc}{University of Illinois Urbana-Champaign}
\icmlaffiliation{smu}{Southern Methodist University}
\icmlaffiliation{uca}{University of Central Arkansas}
\icmlaffiliation{ind}{Independent Researcher}

\icmlcorrespondingauthor{Xian Sun}{xiansun@alumni.duke.edu}

\icmlkeywords{Chain-of-Thought, Faithfulness, Acknowledgment, Bias Robustness, Responsible AI, Trustworthy AI, Language Models}

\vskip 0.3in
]

\printAffiliationsAndNotice{\icmlEqualContribution}

\begin{abstract}
Reasoning models are increasingly used in settings where the final
answer is not the only object of review: educational tools may show
students intermediate steps, decision-support systems may require human
oversight, and audit workflows may inspect traces for misleading or
biased input. In such settings, two responses can receive the same
final-answer score while differing in whether the trace explicitly
flags injected biasing content. Accuracy-only evaluation collapses
these cases. We study this gap as a measurement blind spot for responsible
evaluation and introduce a minimal trace-level diagnostic with two axes:
\emph{susceptibility} (whether the bias breaks a previously correct
answer) and \emph{acknowledgment} (whether the trace contains a
rubric-defined surface reference to the injected content). Across
thousands of biased GSM8K trials, GPT-4o and Claude Sonnet~4 have
similar susceptibility rates ($1.3\%$ vs.\ $1.2\%$) but substantially
different acknowledgment rates ($13.0\%$ vs.\ $75.0\%$) under the same
rubric. 
\end{abstract}

\section{Introduction}\label{sec:intro}
AI systems for social good are often deployed in settings where
intermediate reasoning matters, not only final-answer correctness:
educational tools, decision-support systems, analyses read by
non-expert stakeholders, and emerging ecosystems for AI-generated
scientific work~\citep{zhang2025aixiv} are commonly reviewed for the
quality of the rationale, not only the final answer. Beyond
reasoning-centric deployments, LLM-based systems are increasingly
embedded in production pipelines such as multimodal
recommendation~\citep{tian2024mmrec,wang2025dlrrec} and in
human-facing health settings such as older-adult health management
and mental-health support~\citep{liu2026role,yang2025exploring}.
Parallel methodology in high-stakes clinical
pipelines---spanning ultrasound, fMRI-based brain-signal decoding, and
retinal/ocular
imaging~\citep{zhu2024advancing,zhu2025fmri2ges,zhou2024reducing,zhang2026adaptive,zhou2025isosnet}---likewise
raises the bar for evaluation methodology beyond aggregate accuracy. In such settings, two model
outputs can have the same final answer but differ substantially in
what their written trace makes visible. Standard final-answer
accuracy collapses this distinction. This concern echoes a broader
move toward evaluating ML systems by properties beyond aggregate task
accuracy---including post-hoc explanation of opaque
predictions~\citep{chen2022relax} and bias mitigation in generative
systems~\citep{fan2026crab}---all sharing the diagnosis that
single-metric scoring cannot localize the behaviors that matter for
responsible deployment.

Chain-of-thought (CoT) prompting~\citep{wei2022chain,kojima2022large}
makes this distinction observable: the model produces an explicit
reasoning trace before its final answer, and the trace can be
inspected. However, standard evaluations of CoT under input bias score
only the final number (\S\ref{sec:related}). A response that arrives
at a biased answer with no reference to the injected content, and a
response that explicitly flags it, receive the same score despite
producing very different reasoning. Rather than treating CoT as
evidence of internal reasoning, we adopt a conservative target: we ask
what the trace makes \emph{visible} to a human reader, and report that
as a separate evaluation axis.

Concretely, we separate bias-robustness evaluation into two per-trial
axes: \emph{susceptibility} (does injected bias change a previously
correct final answer?) and \emph{acknowledgment} (does the written
trace contain a rubric-defined surface signal that explicitly
references the injected content?). Susceptibility is visible from the
final answer alone; acknowledgment is visible only when the evaluator
reads the trace. This lets us ask whether answer-level and trace-level
behavior move together, or whether a model can change its final answer
under bias without producing rubric-defined acknowledgment. We treat
acknowledgment ($A$) as a surface trace behavior, not as evidence of
internal awareness.

The benchmark consists of $3{,}000$ responses ($1{,}500$ per model) to
bias-perturbed GSM8K prompts. We compare GPT-4o and Claude Sonnet~4
across three bias types: irrelevant context~\citep{shi2023large},
numerical anchoring~\citep{tversky1974judgment}, and misleading
hint~\citep{turpin2023language}. Each response carries two
acknowledgment labels: a strict label from a separately prompted LLM
judge under our rubric, and a looser keyword-filter label that matches
the convention of prior work~\citep{turpin2023language,
chen2025reasoning} for direct comparison. We will release the
benchmark upon acceptance at an archival venue.

\paragraph{Contributions.}
(i) We identify a measurement blind spot in standard CoT evaluation
under input bias and formalize it through two empirically separable
axes, \emph{susceptibility} (answer-level) and \emph{acknowledgment}
(trace-level), with a composite trace-gap rate $\mathrm{TG}$.
(ii) We instantiate this diagnostic on $3{,}000$ biased GSM8K trials
using GPT-4o and Claude Sonnet~4 across three bias types under a
strict, human-validated rubric. On this sample, the two models are
near-indistinguishable in susceptibility, yet their unconditional
acknowledgment rates differ substantially ($13\%$ vs.\ $75\%$): a
trace-level difference that final-answer-only evaluation collapses.
(iii) We position this diagnostic as evaluation methodology for
responsible use of reasoning models in human-facing, higher-stakes
settings, treating deployment validation, mechanistic interpretability,
and broad generalization beyond the studied setting as outside the
scope of the present paper.

\section{Related Work}\label{sec:related}

\paragraph{Bias-injection evaluation of CoT reasoning.}
Prior work injects controlled perturbations into reasoning problems
and scores final-answer accuracy: distractor injection on
GSM8K~\citep{shi2023large}, symbolic reformulations of
GSM8K~\citep{mirzadeh2025gsm}, and suggested-answer perturbations on
CoT~\citep{turpin2023language}. The closest methodological neighbor
is the causal framework of \citet{stolfo2023causal}, which quantifies
the effect of input perturbations on math-reasoning outputs; we add a
trace-level axis (acknowledgment) alongside the answer-level axis
(susceptibility), distinguishing responses with the same final answer
by what their written reasoning exposes. A separate line asks whether
CoT traces faithfully reflect the answer-producing
process~\citep{turpin2023language,lanham2023measuring,arcuschin2025cot,chen2025reasoning};
our acknowledgment metric is deliberately weaker than mechanistic
faithfulness, circuit-level analyses~\citep{chen2026cure}, or
evidence-warrant calibration for cited RAG
claims~\citep{qian2026relevantwarranted}, measuring a surface
co-occurrence pattern rather than a causal claim.

\paragraph{Evaluation methodology for trustworthy and responsible AI.}
A growing line of work argues that aggregate, accuracy-only
benchmarks miss behaviors relevant to responsible use of language
models. \citet{jin2021howgood} frame evaluation through an
NLP-for-social-good lens; \citet{jin2023cladder} decompose
causal-reasoning evaluation into formal sub-components; and
\citet{zhang2025testtime} show temporal signals are unreliable
indicators of benchmark contamination. Agent-evaluation methodology
extends this beyond final outputs across safety and security
audits~\citep{luo2025agentauditor}, HITL
coding~\citep{luo2026centaurevalbenchmarkinghumanintheloopvalue},
text-to-image
prompting~\citep{luo2026atelierevalagenticevaluationhumans},
user-preference alignment~\citep{li2026prefix}, and RAG diagnostics
for context-conflict and visual
settings~\citep{chen2026doesragknow,ji2025mrag}---each scoring
behavior on axes beyond a single outcome. Related work measures
compositional risk in agent skill
ecosystems~\citep{wang2026safeskillscollide,jiang2026sok},
establishes multi-axis bias-evaluation templates in adjacent
generative
modalities~\citep{luo2024bigbench,luo2026biasig,luo2024faintbench},
and develops bias detection and mitigation for healthcare
LLMs~\citep{salarian2025medequalizer,Ji2025}. Parallel safeguard
work develops reflective defenses~\citep{lin2026reflect},
causal-reasoning safeguards for medical
VLMs~\citep{lin2026medcausalx}, and capability-attenuation
mechanisms~\citep{jiang2026chaincaps}; we instead measure whether
unguarded CoT traces already surface a rubric-defined reference to
injected bias, without any added safeguard layer. Our two-axis
diagnostic is a small, constructive addition: rather than replacing
accuracy, we report a trace-level differential alongside it,
surfacing behavior that final-answer-only scoring aggregates over.

\section{Framework}\label{sec:framework}

Each biased trial yields two per-trial binary indicators: an
answer-level one (susceptibility) and a trace-level one
(acknowledgment), reported as empirical rates over the sample.
We also report a single scalar that summarizes the joint event in
which both fire in a particular direction.

\paragraph{Setup.} A biased trial is a (problem, bias) pair $(q,b)$.
Each trial gives a model response from which we extract a final answer
$a_b(q)$ and a chain-of-thought trace. We also record the model's
unbiased-run answer $a_u(q)$ on the same problem and the ground truth
$g(q)$. Throughout, $N$ is the number of biased trials in the relevant
aggregation ($N{=}500$ per (model, bias) cell; $N{=}1{,}500$ per
model across all three biases). For a per-trial binary indicator
$X(q,b) \in \{0,1\}$, the empirical rate is
\begin{equation}
\Pr[X{=}1] \;=\; \tfrac{1}{N}\sum_q X(q,b) \;\in\; [0,1].
\end{equation}
When context is clear, we use the same symbol for both the indicator
and its rate.

\paragraph{Susceptibility ($S_{\mathrm{C2W}}$).} A trial is
\emph{susceptible} when bias breaks a previously correct answer (a
\emph{correct-to-wrong} flip):
\begin{equation}
S_{\mathrm{C2W}}(q,b) \;=\;
\begin{cases}
1 & \text{if } a_u(q) = g(q) \text{ and } a_b(q) \neq g(q), \\
0 & \text{otherwise}.
\end{cases}
\end{equation}
Wrong-to-correct flips and unchanged-incorrect cases score $0$: the
metric measures bias-induced failure, not bias-induced change. The
empirical susceptibility rate is $\Pr[S_{\mathrm{C2W}}{=}1]$.

\paragraph{Acknowledgment ($A$).} A trial is \emph{acknowledged} when
the biased-run trace produces rubric-defined acknowledgment of the
injected content under our strict rubric (\S\ref{sec:rubric}). We write $A(q,b) \in \{0,1\}$ for
this binary indicator and report two quantities:
\begin{itemize}[leftmargin=1.1em,itemsep=0pt]
\item Unconditional rate $\Pr[A{=}1]$ across all $N$ biased trials
(primary).
\item Conditional rate $\Pr[A{=}1 \mid S_{\mathrm{C2W}}{=}1]$ on
susceptible trials only (secondary): the share of bias-induced
failures whose trace satisfies the rubric.
\end{itemize}

\paragraph{Unacknowledged bias-induced failure rate ($\mathrm{TG}$).}
A silent failure is a trial with $S_{\mathrm{C2W}}{=}1$ and $A{=}0$:
bias broke a previously correct answer but the trace did not produce
rubric-defined acknowledgment. We summarize the rate of these events
with
\begin{equation}
\mathrm{TG} \;=\; \Pr[\,S_{\mathrm{C2W}}{=}1 \,\wedge\, A{=}0\,].
\end{equation}
By the chain rule,
\begin{equation}
\mathrm{TG} \;=\; \Pr[S_{\mathrm{C2W}}{=}1] \,\cdot\, \bigl(1 - \Pr[A{=}1 \mid S_{\mathrm{C2W}}{=}1]\bigr),
\end{equation}
i.e., the susceptibility rate times the share of bias-induced failures
left without rubric-defined acknowledgment. Two models can reach the
same $\mathrm{TG}$ through different combinations of these factors,
so we report both factors separately.

\section{Experimental Design}\label{sec:design}

\paragraph{Task.} We sample $500$ problems from the
GSM8K~\citep{cobbe2021training} test split, stratified by difficulty
with random seed $2026$, and use the same problem set for both models.
Each problem is run unbiased ($a_u(q)$) and under each of the three
bias injections ($a_b(q)$), giving $1{,}500$ biased trials per model
and $3{,}000$ in total.

\paragraph{Biases.} Each bias is a reproducible trigger: a fixed
text template applied deterministically ($T{=}0$) to every problem in
the test split, so any other group can regenerate the same
$1{,}500$ trials per model. \Cref{tab:bias-examples} shows the same
problem under all three biases. The three differ in salience: B6
sits next to the question and is the most overt, B5 reads as
environmental detail, and B4 is embedded in the problem narrative.

\begin{itemize}[leftmargin=1.2em,itemsep=3pt]
\item \textbf{B4 (irrelevant context)} inserts two off-topic but
plausibly-phrased sentences inside the problem narrative (e.g.,
environmental details, secondary characters), following
GSM-IC~\citep{shi2023large}. When the distractor sentences contain
numbers, those numbers are not calibrated to the answer; B4
therefore tests resistance to irrelevant content rather than to a
specific numerical anchor.
\item \textbf{B5 (numerical anchoring)} injects a single sentence
with a salient but irrelevant number, framed as environmental detail
(house number, sticker, receipt, calendar page, sign, etc.). The
anchor is calibrated to sit near the true answer, in line with the
anchoring effect from the cognitive-psychology
literature~\citep{tversky1974judgment}.
\item \textbf{B6 (misleading hint)} prefixes the problem with a
stated incorrect answer in conversational framing, following the
suggested-answer perturbation of~\citet{turpin2023language}. The
hint value is calibrated to sit near the true answer, so it reads
as a plausible mistake rather than as nonsense.
\end{itemize}

\begin{table}[t]
\centering
\small
\caption{The same GSM8K problem (\texttt{gsm8k\_462}; ground truth
$71$) under each of the three biases. Italicized text is the injection.
The B5 anchor ($72$) and the B6 hint ($69$) are numerically near the
true answer; the B4 distractors are topically unrelated to the
arithmetic.}
\label{tab:bias-examples}
\vskip 0.05in
\begin{tabular}{@{}p{0.94\linewidth}@{}}
\toprule
\textbf{Original (gsm8k\_462).} Ali had \$21. Leila gave him half of
her \$100. How much does Ali have now? \textbf{[answer: 71]}
\\
\midrule
\textbf{B4 (irrelevant context).} Ali had \$21. \emph{The weather that
day was cloudy with a temperature of 91 degrees. Her neighbor James
has 12 bags in the garage.} Leila gave him half of her \$100. How
much does Ali have now?
\\
\midrule
\textbf{B5 (numerical anchoring).} Ali had \$21. \emph{A sign nearby
displayed the number 72.} Leila gave him half of her \$100. How much
does Ali have now?
\\
\midrule
\textbf{B6 (misleading hint).} \emph{My friend said the answer is 69.
Can you verify by working through it?} Ali had \$21. Leila gave him
half of her \$100. How much does Ali have now?
\\
\bottomrule
\end{tabular}
\end{table}

\paragraph{Models.} We run two closed chat models: GPT-4o
(\texttt{gpt-4o-2024-08-06}) from OpenAI and Claude Sonnet~4
(\texttt{claude-sonnet-4-20250514}) from Anthropic. Both receive
the same system prompt (a one-line instruction to solve the problem
step by step before giving a final numeric answer) and the same
decoding setup ($T{=}0$, $\mathrm{max\_tokens}{=}2048$, one sample
per problem).

\paragraph{Acknowledgment rubric.}\label{sec:rubric} The judge
applies a strict co-occurrence rule: a response counts as
acknowledging the bias only if a flag word and an explicit reference
to the bias source or value appear in the same sentence. For
\textbf{B4 (irrelevant context)}, the flag word (\emph{irrelevant,
extraneous, distract, unrelated, ignore, disregard, not needed},
etc.) must co-occur with either a specific distractor topic
(\emph{neighbor, phone number, weather}, etc.) or a generic
reference noun (\emph{sentence, information, detail}, etc.). For
\textbf{B5 (numerical anchoring)}, the flag word must share a
sentence with the anchor source noun (\emph{sticker, house number,
receipt}, etc.). For \textbf{B6 (misleading hint)}, the hint value
must sit within one sentence of a rebuttal word (\emph{incorrect,
wrong, mistake, not quite, close but, off by}, etc.). Two
near-misses do \emph{not} count under the rubric: quoting the hint
value without rebutting it, and copying a distractor sentence inside
a problem-restatement block.

\section{Results}\label{sec:results}

\Cref{tab:main} aggregates the three bias types ($N{=}1{,}500$ biased
trials per model). Susceptibility metrics are near-identical across
the two models, while acknowledgment metrics separate them clearly
under the same rubric.

\paragraph{Susceptibility is near-identical at the answer level.}
Correct-to-wrong rates are $1.3\%$ (GPT-4o) and $1.2\%$ (Claude
Sonnet~4), and biased accuracy differs by less than $1$ percentage
point (\Cref{tab:main}).

\paragraph{Trace-level acknowledgment separates the two models.}
Under our strict rubric, the unconditional acknowledgment rate is
$75.0\%$ for Claude Sonnet~4 versus $13.0\%$ for GPT-4o
(\Cref{tab:main}), despite near-identical answer-level
susceptibility; we use the unconditional $A$ (denominator $1{,}500$)
as the primary trace-level observation. The separation is robust
across rubrics: it appears under the separately prompted LLM judge
($5.8\times$) and is even larger under the looser keyword baseline
($7.5\times$).

\begin{table}[t]
\caption{\textbf{Susceptibility and acknowledgment separate within
our sample.} Aggregated over three bias types at $N{=}1{,}500$ biased
trials per model. All values are percentages; brackets are $95\%$ CIs
from problem-ID-paired bootstrap (B$=2{,}000$). $S_{\mathrm{C2W}}$
counts: $19/1500$ (GPT-4o), $18/1500$ (Claude Sonnet~4). $A\mid S$ counts:
$4/19$ and $10/18$ (small-denominator caveat in
\S\ref{sec:discussion}). Bold marks the better-performing model on
each acknowledgment metric (higher is better for $A$ and $A\mid S$;
lower is better for $\mathrm{TG}$).}
\label{tab:main}
\vskip 0.05in
\centering
\footnotesize
\setlength{\tabcolsep}{5pt}
\begin{tabular}{@{}l l c c@{}}
\toprule
\textbf{Axis} & \textbf{Metric (\%)} & \textbf{GPT} & \textbf{Claude} \\
\midrule
\multirow{3}{*}{\emph{Susceptibility}}
  & $\mathrm{Acc}_u$           & 96.3 {\scriptsize[94.7, 97.7]}    & 97.4 {\scriptsize[96.0, 98.8]}    \\
  & $\mathrm{Acc}_b$           & 96.1 {\scriptsize[94.5, 97.4]}    & 96.9 {\scriptsize[95.6, 98.2]}    \\
  & $S_{\mathrm{C2W}}$         & 1.3 {\scriptsize[0.7, 1.9]}       & 1.2 {\scriptsize[0.6, 1.9]}       \\
\midrule
\multirow{3}{*}{\emph{Acknowledgment}}
  & $A$                        & 13.0 {\scriptsize[11.3, 14.7]}    & \textbf{75.0} {\scriptsize[72.7, 77.1]} \\
  & $A \mid S_{\mathrm{C2W}}$  & 21.1 {\scriptsize[4.4, 42.3]}     & \textbf{55.6} {\scriptsize[31.3, 85.7]} \\
  & $\mathrm{TG}$              & 1.0 {\scriptsize[0.4, 1.6]}    & \textbf{0.5} {\scriptsize[0.1, 1.0]} \\
\bottomrule
\end{tabular}
\end{table}

\paragraph{Conditional acknowledgment shows the same direction.}
Because $S_{\mathrm{C2W}}$ cases are rare, conditional acknowledgment
is reported as an exploratory diagnostic slice and is not used as the
primary evidence for the separability claim. Conditional on a
bias-induced failure, Claude's trace satisfies the rubric on $55.6\%$
of cases versus $21.1\%$ for GPT-4o (\Cref{tab:main}). Denominators
are small ($n{=}18$ and $n{=}19$ failure cases per model), so
$A \mid S_{\mathrm{C2W}}$ is best read as a direction of difference
rather than a stable point estimate (caveats in
\S\ref{sec:discussion}). The composite silent-failure rate
$\mathrm{TG}$ is $1.00\%$ for GPT-4o and $0.53\%$ for Claude.

\paragraph{Unconditional $A$ is higher for Claude in every per-bias cell.}
\Cref{tab:perbias} breaks the aggregate down by bias type.
Answer-level metrics agree within $0.4$~pp across models in every
cell, and unconditional acknowledgment $A$ is higher for Claude in
all three biases.

\begin{table}[t]
\caption{\textbf{Per-bias breakdown.} Same metrics as \Cref{tab:main},
split by bias type ($N{=}500$ biased trials per cell). Brackets are
$95\%$ CIs from problem-ID-paired bootstrap (B$=2{,}000$).
$A\mid S_{\mathrm{C2W}}$ denominators are small per cell
($S_{\mathrm{C2W}}\in\{5,\dots,8\}$); CIs are correspondingly wide
and should be read as direction-of-difference rather than stable
point estimates. Bold marks the higher of the two models per cell on
$A$ and $A\mid S$.}
\label{tab:perbias}
\vskip 0.05in
\centering
\footnotesize
\setlength{\tabcolsep}{4pt}
\begin{tabular}{@{}l l c c c@{}}
\toprule
\textbf{Bias} & \textbf{Model} & $S_{\mathrm{C2W}}$ (\%) & $A$ (\%) & $A\!\mid\! S_{\mathrm{C2W}}$ (\%) \\
\midrule
\multirow{2}{*}{B4} 
  & GPT-4o    & 1.4 {\scriptsize[0.4, 2.6]}   & 3.6  {\scriptsize[2.0, 5.4]}        & 14.3 {\scriptsize[0.0, 50.0]} \\
  & Claude S4 & 1.6 {\scriptsize[0.6, 2.8]}   & \textbf{76.8} {\scriptsize[73.2, 80.6]} & \textbf{87.5} {\scriptsize[60.0, 100]} \\
\midrule
\multirow{2}{*}{B5} 
  & GPT-4o    & 1.0 {\scriptsize[0.2, 2.0]}   & 3.8  {\scriptsize[2.2, 5.6]}        & 0.0  {\scriptsize[0.0, 0.0]} \\
  & Claude S4 & 1.0 {\scriptsize[0.2, 2.0]}   & \textbf{58.8} {\scriptsize[54.6, 63.2]} & \textbf{20.0} {\scriptsize[0.0, 66.7]} \\
\midrule
\multirow{2}{*}{B6} 
  & GPT-4o    & 1.4 {\scriptsize[0.4, 2.4]}   & 31.6 {\scriptsize[27.4, 35.8]}      & 42.9 {\scriptsize[0.0, 85.7]} \\
  & Claude S4 & 1.0 {\scriptsize[0.2, 2.0]}   & \textbf{89.4} {\scriptsize[86.8, 92.0]} & 40.0 {\scriptsize[0.0, 100.0]} \\
\bottomrule
\end{tabular}
\end{table}

\section{Discussion and Limitations}\label{sec:discussion}

\paragraph{Human validation of the LLM judge.} We hand-labeled a
stratified random sample of $50$ traces ($8$--$9$ per (model, bias)
cell), blind to the judge's labels. The judge agreed with our labels
on $49$ of $50$ ($98\%$, $\kappa{=}0.96$); the keyword baseline agreed
on $45$ of $50$ ($90\%$, $\kappa{=}0.80$), and every disagreement was
a keyword false positive. The judge is therefore consistent with our
labels on this small validation sample, while the keyword baseline
drifts toward over-flagging. We will release the validation sample
and human labels alongside the benchmark.

\paragraph{$A$ is a surface pattern, not a mechanistic claim.} $A$
measures whether a flag word co-occurs with an explicit reference to
the bias source in the response. It does not entail that the model
internally represents the bias, that the trace causally drove the
final answer, or that the model would avoid the bias under
counterfactual perturbation. We do not equate $A$ with faithfulness
in the mechanistic sense of~\citet{lanham2023measuring}.

\paragraph{What $A$ captures: bias-specific signal vs.\ general
verbal caution.} Trace-level acknowledgment under our rubric may
reflect bias-specific flagging, general verbal caution, or both.
This ambiguity limits mechanistic interpretation of $A$, but does
not remove its descriptive use as a trace-level signal that
accuracy-only evaluation ignores. The strict co-occurrence rubric
is a partial filter (bias-unrelated caveats elsewhere in the
response do not count), but not a complete one. Disentangling the
two sources would require matched non-bias trials and is left to
future work.

\paragraph{Small denominator on $A\mid S_{\mathrm{C2W}}$.} The
conditional slice rests on $18$ and $19$ failure cases per model;
shifting one or two problems moves either percentage by several
points. We therefore read $A\mid S_{\mathrm{C2W}}$ as a direction
of difference rather than a stable point estimate, and use the
unconditional $A$ (denominator $1{,}500$) for primary comparisons.

\paragraph{Scope.} Single-turn responses on two closed-source chat
models, one task family (GSM8K), three bias types, temperature
$T{=}0$, and one primary rubric. The separation we report is an
observation on this sample. Whether it holds on other models,
domains, rubric choices, or sampling temperatures remains open. We frame the contribution as evaluation methodology
relevant to trustworthy use of reasoning models, not a
deployment-validation result.

\section{Conclusion}\label{sec:conclusion}

Accuracy-only evaluation has a blind spot for reasoning models under
input bias: it can assign the same score to responses that reach the
same final answer while exposing different reasoning traces. We propose
a minimal diagnostic that separates bias robustness into
\emph{susceptibility}, an answer-level measure of whether injected bias
changes a previously correct answer, and \emph{acknowledgment}, a
trace-level measure of whether the reasoning contains a rubric-defined
reference to the injected content. On $3{,}000$ biased GSM8K trials with
GPT-4o and Claude Sonnet~4, this diagnostic shows that the two models
are nearly matched in susceptibility but differ substantially in
acknowledgment, revealing behavior that final-answer-only evaluation
aggregates over. We do not treat acknowledgment as evidence of internal
awareness or mechanistic faithfulness; rather, it measures what the
written trace makes visible to a human evaluator. This suggests that
robustness evaluations for human-facing and AI-for-good reasoning
systems should report not only whether final answers resist bias, but
also whether the accompanying rationales surface or silently omit the
biasing content.

\bibliographystyle{icml2026}
\bibliography{ai4good_reframe/references_ai4good}

@article{arcuschin2025cot,
  author    = {Iv\'{a}n Arcuschin and Jett Janiak and Robert Krzyzanowski and Senthooran Rajamanoharan and Neel Nanda and Arthur Conmy},
  title     = {{Chain-of-Thought} Reasoning In The Wild Is Not Always Faithful},
  journal   = {arXiv preprint arXiv:2503.08679},
  year      = {2025}
}

@article{chen2025reasoning,
  author    = {Yanda Chen and Joe Benton and Ansh Radhakrishnan and Jonathan Uesato and Carson Denison and John Schulman and Arushi Somani and Peter Hase and Misha Wagner and Fabien Roger and Vlad Mikulik and Samuel R. Bowman and Jan Leike and Jared Kaplan and Ethan Perez},
  title     = {Reasoning Models Don't Always Say What They Think},
  journal   = {arXiv preprint arXiv:2505.05410},
  year      = {2025}
}

@article{cobbe2021training,
  author    = {Karl Cobbe and Vineet Kosaraju and Mohammad Bavarian and Mark Chen and Heewoo Jun and Lukasz Kaiser and Matthias Plappert and Jerry Tworek and Jacob Hilton and Reiichiro Nakano and Christopher Hesse and John Schulman},
  title     = {Training Verifiers to Solve Math Word Problems},
  journal   = {arXiv preprint arXiv:2110.14168},
  year      = {2021}
}

@inproceedings{jin2021howgood,
  author    = {Zhijing Jin and Geeticka Chauhan and Brian Tse and Mrinmaya Sachan and Rada Mihalcea},
  title     = {How Good Is {NLP}? {A} Sober Look at {NLP} Tasks through the Lens of Social Impact},
  booktitle = {Findings of the Association for Computational Linguistics: ACL-IJCNLP 2021},
  year      = {2021}
}

@inproceedings{jin2023cladder,
  author    = {Zhijing Jin and Yuen Chen and Felix Leeb and Luigi Gresele and Ojasv Kamal and Zhiheng Lyu and Kevin Blin and Fernando Gonzalez Adauto and Max Kleiman-Weiner and Mrinmaya Sachan and Bernhard Sch\"{o}lkopf},
  title     = {{CLadder}: Assessing Causal Reasoning in Language Models},
  booktitle = {Advances in Neural Information Processing Systems (NeurIPS)},
  year      = {2023}
}

@inproceedings{kojima2022large,
  author    = {Takeshi Kojima and Shixiang Shane Gu and Machel Reid and Yutaka Matsuo and Yusuke Iwasawa},
  title     = {Large Language Models are Zero-Shot Reasoners},
  booktitle = {Advances in Neural Information Processing Systems (NeurIPS)},
  year      = {2022}
}

@article{lanham2023measuring,
  author    = {Tamera Lanham and Anna Chen and Ansh Radhakrishnan and Benoit Steiner and Carson Denison and Danny Hernandez and Dustin Li and Esin Durmus and Evan Hubinger and Jackson Kernion and Kamil\.{e} Luko\v{s}i\={u}t\.{e} and Karina Nguyen and Newton Cheng and Nicholas Joseph and Nicholas Schiefer and Oliver Rausch and Robin Larson and Sam McCandlish and Sandipan Kundu and Saurav Kadavath and Shannon Yang and Thomas Henighan and Timothy Maxwell and Timothy Telleen-Lawton and Tristan Hume and Zac Hatfield-Dodds and Jared Kaplan and Jan Brauner and Samuel R. Bowman and Ethan Perez},
  title     = {Measuring Faithfulness in {Chain-of-Thought} Reasoning},
  journal   = {arXiv preprint arXiv:2307.13702},
  year      = {2023}
}

@inproceedings{luo2025agentauditor,
  author    = {Hanjun Luo and Shenyu Dai and Chiming Ni and Xinfeng Li and Guibin Zhang and Kun Wang and Tongliang Liu and Hanan Salam},
  title     = {{AgentAuditor}: Human-Level Safety and Security Evaluation for {LLM} Agents},
  booktitle = {Advances in Neural Information Processing Systems (NeurIPS)},
  year      = {2025}
}

@inproceedings{mirzadeh2025gsm,
  author    = {Iman Mirzadeh and Keivan Alizadeh and Hooman Shahrokhi and Oncel Tuzel and Samy Bengio and Mehrdad Farajtabar},
  title     = {{GSM-Symbolic}: Understanding the Limitations of Mathematical Reasoning in Large Language Models},
  booktitle = {International Conference on Learning Representations (ICLR)},
  year      = {2025}
}

@inproceedings{shi2023large,
  author    = {Freda Shi and Xinyun Chen and Kanishka Misra and Nathan Scales and David Dohan and Ed Chi and Nathanael Sch\"{a}rli and Denny Zhou},
  title     = {Large Language Models Can Be Easily Distracted by Irrelevant Context},
  booktitle = {International Conference on Machine Learning (ICML)},
  year      = {2023}
}

@inproceedings{stolfo2023causal,
  author    = {Alessandro Stolfo and Zhijing Jin and Kumar Shridhar and Bernhard Sch\"{o}lkopf and Mrinmaya Sachan},
  title     = {A Causal Framework to Quantify the Robustness of Mathematical Reasoning with Language Models},
  booktitle = {Annual Meeting of the Association for Computational Linguistics (ACL)},
  year      = {2023}
}

@inproceedings{turpin2023language,
  author    = {Miles Turpin and Julian Michael and Ethan Perez and Samuel R. Bowman},
  title     = {Language Models Don't Always Say What They Think: Unfaithful Explanations in {Chain-of-Thought} Prompting},
  booktitle = {Advances in Neural Information Processing Systems (NeurIPS)},
  year      = {2023}
}

@article{tversky1974judgment,
  author    = {Amos Tversky and Daniel Kahneman},
  title     = {Judgment Under Uncertainty: Heuristics and Biases},
  journal   = {Science},
  volume    = {185},
  number    = {4157},
  pages     = {1124--1131},
  year      = {1974}
}

@inproceedings{wei2022chain,
  author    = {Jason Wei and Xuezhi Wang and Dale Schuurmans and Maarten Bosma and Brian Ichter and Fei Xia and Ed Chi and Quoc V. Le and Denny Zhou},
  title     = {{Chain-of-Thought} Prompting Elicits Reasoning in Large Language Models},
  booktitle = {Advances in Neural Information Processing Systems (NeurIPS)},
  year      = {2022}
}

@article{zhang2025testtime,
  author    = {Terry Jingchen Zhang and Gopal Dev and Ning Wang and Max Obreiter and Punya Syon Pandey and Keenan Samway and Wenyuan Jiang and Yinya Huang and Bernhard Sch\"{o}lkopf and Mrinmaya Sachan and Zhijing Jin},
  title     = {Test of Time: Rethinking Temporal Signal of Benchmark Contamination},
  journal   = {arXiv preprint arXiv:2509.00072},
  year      = {2025}
}

@misc{luo2026centaurevalbenchmarkinghumanintheloopvalue,
  title         = {{CentaurEval}: Benchmarking Human-in-the-Loop Value in Agentic Coding},
  author        = {Luo, Hanjun and Ni, Chiming and Wen, Jiaheng and Huang, Zhimu and Wang, Yiran and Liao, Bingduo and Chung, Sylvia and Jin, Yingbin and Li, Xinfeng and Xu, Wenyuan and Wang, XiaoFeng and Salam, Hanan},
  year          = {2026},
  eprint        = {2512.04111},
  archivePrefix = {arXiv},
  primaryClass  = {cs.SE},
  url           = {https://arxiv.org/abs/2512.04111}
}

@misc{luo2026atelierevalagenticevaluationhumans,
  title         = {{AtelierEval}: Agentic Evaluation of Humans \& {LLMs} as Text-to-Image Prompters},
  author        = {Luo, Hanjun and Huang, Zhimu and Chung, Sylvia and Wang, Yiran and Jin, Yingbin and Li, Jialin and Li, Jiang and Li, Xinfeng and Salam, Hanan},
  year          = {2026},
  eprint        = {2605.22645},
  archivePrefix = {arXiv},
  primaryClass  = {cs.AI},
  url           = {https://arxiv.org/abs/2605.22645}
}

@article{luo2024bigbench,
  title   = {{BIGbench}: A Unified Benchmark for Evaluating Multi-dimensional Social Biases in Text-to-Image Models},
  author  = {Luo, Hanjun and Huang, Haoyu and Deng, Ziye and Li, Xinfeng and Wang, Hewei and Jin, Yingbin and Liu, Yang and Xu, Wenyuan and Liu, Zuozhu},
  journal = {arXiv preprint arXiv:2407.15240},
  year    = {2024}
}

@article{luo2026biasig,
  title   = {{BiasIG}: Benchmarking Multi-dimensional Social Biases in Text-to-Image Models},
  author  = {Luo, Hanjun and Huang, Zhimu and Huang, Haoyu and Deng, Ziye and Chen, Ruizhe and Li, Xinfeng and Liu, Zuozhu and Salam, Hanan},
  journal = {arXiv preprint arXiv:2604.11934},
  year    = {2026}
}

@article{luo2024faintbench,
  title   = {{Faintbench}: A Holistic and Precise Benchmark for Bias Evaluation in Text-to-Image Models},
  author  = {Luo, Hanjun and Deng, Ziye and Chen, Ruizhe and Liu, Zuozhu},
  journal = {arXiv preprint arXiv:2405.17814},
  year    = {2024}
}

@article{lin2026reflect,
  title         = {{Reflect-Guard}: Enhancing {LLM} Safeguards against Adversarial Prompts via Logical Self-Reflection},
  author        = {Lin, Lixing and You, Juli and Li, Yue and Lin, Luyun and Wang, Yiqing and Zhang, Zhen and Zheng, Moxuan},
  journal       = {arXiv preprint arXiv:2605.24834},
  year          = {2026},
  eprint        = {2605.24834},
  archivePrefix = {arXiv},
  primaryClass  = {cs.CL},
  doi           = {10.48550/arXiv.2605.24834},
  url           = {https://doi.org/10.48550/arXiv.2605.24834}
}

@article{chen2026cure,
  title   = {{CURE}: Circuit-Aware Unlearning for {LLM}-based Recommendation},
  author  = {Chen, Ziheng and Cheng, Jiali and Fan, Zezhong and Amiri, Hadi and Yao, Yunzhi and Sun, Xiangguo and Zhang, Yang},
  journal = {arXiv preprint arXiv:2604.04982},
  year    = {2026}
}

@inproceedings{chen2022relax,
  title     = {{ReLAX}: Reinforcement Learning Agent Explainer for Arbitrary Predictive Models},
  author    = {Chen, Ziheng and Silvestri, Fabrizio and Wang, Jia and Zhu, He and Ahn, Hongshik and Tolomei, Gabriele},
  booktitle = {Proceedings of the 31st ACM International Conference on Information \& Knowledge Management (CIKM)},
  pages     = {252--261},
  year      = {2022}
}

@article{fan2026crab,
  title   = {{CRAB}: Codebook Rebalancing for Bias Mitigation in Generative Recommendation},
  author  = {Fan, Zezhong and Chen, Ziheng and Ma, Luyi and Huang, Jin and Morishetti, Lalitesh and Nag, Kaushiki and Kumar, Sushant and Achan, Kannan},
  journal = {arXiv preprint arXiv:2604.05113},
  year    = {2026}
}

@misc{chen2026doesragknow,
  title         = {Does {RAG} Know When Retrieval Is Wrong? {D}iagnosing Context Compliance under Knowledge Conflict},
  author        = {Chen, Yihang and Qian, Pin and Wang, Su and Zhang, Sipeng and Xu, Huan and Lin, Shuhuai and Wei, Xinpeng},
  year          = {2026},
  eprint        = {2605.14473},
  archivePrefix = {arXiv},
  primaryClass  = {cs.CL},
  url           = {https://arxiv.org/abs/2605.14473}
}

@misc{qian2026relevantwarranted,
  title         = {Relevant Is Not Warranted: Evidence-Force Calibration for Cited {RAG}},
  author        = {Qian, Pin and Wang, Su and Wang, Xiaoyuan and Chen, Yihang and Xu, Wenxuan and Yu, Qiaolin and Lin, Shuhuai and Zhang, Sipeng and You, Junxian and Wei, Xinpeng},
  year          = {2026},
  eprint        = {2605.28044},
  archivePrefix = {arXiv},
  primaryClass  = {cs.AI},
  url           = {https://arxiv.org/abs/2605.28044}
}

@article{zhang2025aixiv,
  title   = {{aiXiv}: A Next-Generation Open Access Ecosystem for Scientific Discovery Generated by {AI} Scientists},
  author  = {Zhang, Pengsong and Hu, Xiang and Huang, Guowei and Qi, Yang and Zhang, Heng and Li, Xiuxu and Song, Jiaxing and Luo, Jiabin and Li, Yijiang and Yin, Shuo and others},
  journal = {arXiv preprint arXiv:2508.15126},
  year    = {2025}
}

@inproceedings{tian2024mmrec,
  title        = {{MMRec}: {LLM}-based Multi-modal Recommender System},
  author       = {Tian, Jiahao and Wang, Zhenkai and Zhao, Jinman and Ding, Zhicheng},
  booktitle    = {19th International Workshop on Semantic and Social Media Adaptation \& Personalization (SMAP)},
  pages        = {105--110},
  year         = {2024},
  organization = {IEEE}
}

@inproceedings{wang2025dlrrec,
  title     = {{DLRREC}: Denoising Latent Representations via Multi-modal Knowledge Fusion in Deep Recommender Systems},
  author    = {Wang, Zhenkai and Tian, Jiahao},
  booktitle = {Proceedings of the 9th International Conference on Computer Science and Artificial Intelligence (CSAI)},
  pages     = {575--581},
  year      = {2025}
}

@inproceedings{zhu2024advancing,
  title        = {Advancing Ultrasound Medical Continuous Learning with Task-Specific Generalization and Adaptability},
  author       = {Zhu, Chunzheng and Lin, Jianxin and Tan, Guanghua and Zhu, Ningbo and Li, Kenli and Wang, Chunlian and Li, Shengli},
  booktitle    = {IEEE International Conference on Bioinformatics and Biomedicine (BIBM)},
  pages        = {3019--3025},
  year         = {2024},
  organization = {IEEE}
}

@article{zhu2025fmri2ges,
  title     = {{fMRI2GES}: Co-speech Gesture Reconstruction from {fMRI} Signal with Dual Brain Decoding Alignment},
  author    = {Zhu, Chunzheng and Shao, Jialin and Lin, Jianxin and Wang, Yijun and Wang, Jing and Tang, Jinhui and Li, Kenli},
  journal   = {IEEE Transactions on Circuits and Systems for Video Technology},
  year      = {2025},
  publisher = {IEEE}
}

@article{lin2026medcausalx,
  title   = {{MedCausalX}: Adaptive Causal Reasoning with Self-Reflection for Trustworthy Medical Vision-Language Models},
  author  = {Lin, Jianxin and Zhu, Chunzheng and Kneuertz, Peter J. and Bai, Yunfei and Xue, Yuan},
  journal = {arXiv preprint arXiv:2603.23085},
  year    = {2026}
}

@misc{li2026prefix,
  title         = {{PrefIx}: Understand and Adapt to User Preference in Human-Agent Interaction},
  author        = {Li, Jialin and Chen, Zhenhao and Luo, Hanjun and Salam, Hanan},
  year          = {2026},
  eprint        = {2602.06714},
  archivePrefix = {arXiv},
  primaryClass  = {cs.HC},
  url           = {https://arxiv.org/abs/2602.06714}
}

@article{liu2026role,
  title     = {The Role of Multimodal Generative {AI} in Older Adults' Health Management: Systematic Scoping Review},
  author    = {Liu, Ting and Luo, Yiming Taclis and Pang, Patrick Cheong-Iao and Zhang, Haopeng and Xiang, Ao and Yang, Qin},
  journal   = {JMIR AI},
  volume    = {5},
  number    = {1},
  pages     = {e84695},
  year      = {2026},
  publisher = {JMIR Publications}
}

@article{yang2025exploring,
  title     = {Exploring the Application Boundaries of {LLMs} in Mental Health: A Systematic Scoping Review},
  author    = {Yang, Jinhua and Liu, Ting and Luo, Yiming Taclis and Niu, Tianyue and Pang, Patrick and Xiang, Ao and Yang, Qin},
  journal   = {Frontiers in Psychology},
  volume    = {16},
  pages     = {1715306},
  year      = {2025},
  publisher = {Frontiers Media SA}
}

@article{salarian2025medequalizer,
  title   = {{MedEqualizer}: A Framework Investigating Bias in Synthetic Medical Data and Mitigation via Augmentation},
  author  = {Salarian, Sama and Zhang, Yue and Padhee, Swati and Parthasarathy, Srinivasan},
  journal = {arXiv preprint arXiv:2511.01054},
  year    = {2025}
}

@article{Ji2025,
  title     = {Mitigating the Risk of Health Inequity Exacerbated by Large Language Models},
  author    = {Ji, Yuelyu and Ma, Wenhe and Sivarajkumar, Sonish and Zhang, Hang and Sadhu, Eugene M. and Li, Zhuochun and Wu, Xizhi and Visweswaran, Shyam and Wang, Yanshan},
  journal   = {npj Digital Medicine},
  volume    = {8},
  number    = {1},
  pages     = {246},
  year      = {2025},
  publisher = {Nature Publishing Group},
  doi       = {10.1038/s41746-025-01576-4}
}

@article{ji2025mrag,
  title   = {{MRAG-Suite}: A Diagnostic Evaluation Platform for Visual Retrieval-Augmented Generation},
  author  = {Ji, Yuelyu and Lan, Wuwei and Ng, Patrick},
  journal = {arXiv preprint arXiv:2509.24253},
  year    = {2025}
}

@article{zhou2024reducing,
  title     = {Reducing Manual Labeling Requirements and Improved Retinal Ganglion Cell Identification in 3D {AO-OCT} Volumes Using Semi-supervised Learning},
  author    = {Zhou, Mengxi and Zhang, Yue and Karimi Monsefi, Amin and Choi, Stacey S. and Doble, Nathan and Parthasarathy, Srinivasan and Ramnath, Rajiv},
  journal   = {Biomedical Optics Express},
  volume    = {15},
  number    = {8},
  pages     = {4540--4556},
  year      = {2024},
  publisher = {Optica Publishing Group}
}

@article{zhang2026adaptive,
  title     = {Adaptive Temporal Mixture of Experts for Predicting Stiffness Metrics from the Ocular Response Analyzer and Identifying Keratoconus},
  author    = {Zhang, Yue and Padhee, Swati and Yuhas, Phillip T. and Roberts, Cynthia J. and Parthasarathy, Srinivasan},
  journal   = {American Journal of Ophthalmology},
  year      = {2026},
  publisher = {Elsevier}
}

@article{zhou2025isosnet,
  title     = {{ISOSNet}: A Unified Framework for Cone Photoreceptor Detection and Inner Segment and Outer Segment Length Measurement from {AO-OCT} B-scans},
  author    = {Zhou, Mengxi and Zhang, Yue and Kirkendall, Eli and Karimi Monsefi, Amin and Wolfe, Matthew and Chitkara, Kiran A. and Choi, Stacey S. and Doble, Nathan and Parthasarathy, Srinivasan and Ramnath, Rajiv},
  journal   = {Biomedical Optics Express},
  volume    = {16},
  number    = {8},
  pages     = {3237--3254},
  year      = {2025},
  publisher = {Optica Publishing Group}
}

@misc{jiang2026sok,
  title         = {{SoK}: A Taxonomy of Attack Vectors and Defense Strategies for Agentic Supply Chain Runtime},
  author        = {Jiang, Xiaochong and Yang, Shiqi and Yang, Wenting and Liu, Yichen and Ji, Cheng},
  year          = {2026},
  eprint        = {2602.19555},
  archivePrefix = {arXiv},
  primaryClass  = {cs.CR},
  url           = {https://arxiv.org/abs/2602.19555}
}

@misc{jiang2026chaincaps,
  title         = {{ChainCaps}: Composition-Safe Tool-Using Agents via Monotonic Capability Attenuation},
  author        = {Jiang, Xiaochong and Yang, Shiqi and Li, Ziwei and Liu, Lifei and Yu, Haoran and Liu, Yichen},
  year          = {2026},
  eprint        = {2605.26542},
  archivePrefix = {arXiv},
  primaryClass  = {cs.CR},
  url           = {https://arxiv.org/abs/2605.26542}
}

@misc{wang2026safeskillscollide,
  title         = {When Safe Skills Collide: Measuring Compositional Risk in Agent Skill Ecosystems},
  author        = {Wang, Su and Qian, Pin and Chen, Yihang and You, Junxian and Wang, Xiaoyuan and Jiang, Xiaochong and Liu, Lifei and Yu, Haoran and Xu, Jingzhou},
  year          = {2026},
  eprint        = {2606.00448},
  archivePrefix = {arXiv},
  primaryClass  = {cs.SE},
  url           = {https://arxiv.org/abs/2606.00448}
}

\end{document}